\documentclass[sigconf]{acmart}
\usepackage{siunitx}

\usepackage{textgreek}
 \usepackage{array,multirow,graphicx}
\usepackage{threeparttable}
\def\BibTeX{{\rm B\kern-.05em{\sc i\kern-.025em b}\kern-.08em
    T\kern-.1667em\lower.7ex\hbox{E}\kern-.125emX}}
\usepackage{multirow}
\usepackage[caption=false]{subfig}
\usepackage{tabu}
\usepackage{color, colortbl}
\newcommand{\thickhline}{%
    \noalign {\ifnum 0=`}\fi \hrule height 1pt
    \futurelet \reserved@a \@xhline
}
\usepackage[linesnumbered,ruled,noend]{algorithm2e}
\usepackage{xcolor}

\SetCommentSty{mycommfont}    
\usepackage{pifont}
\usepackage{multirow}
\usepackage{hyperref}

\DeclareUnicodeCharacter{2212}{-}
\AtBeginDocument{%
  \providecommand\BibTeX{{%
    \normalfont B\kern-0.5em{\scshape i\kern-0.25em b}\kern-0.8em\TeX}}}

\def\BibTeX{{\rm B\kern-.05em{\sc i\kern-.025em b}\kern-.08em
    T\kern-.1667em\lower.7ex\hbox{E}\kern-.125emX}}

\acmConference[X]{X}{X}

\begin{document}

\title{A Coupled Neural Circuit Design for Guillain-Barre Syndrome}


\author{Oguzhan Derebasi}
\authornotemark[1]
\email{e170501019@stud.tau.edu.tr}
\affiliation{%
  \institution{Turkish-German University}
  \state{Istanbul}
  \country{Turkey}
  \postcode{34000}
}

\author{Murat Isik}
\email{mci38@drexel.edu}
\authornote{Authors contributed equally to this research.}
\affiliation{%
  \institution{Drexel University}
  \city{Philadelphia}
  \country{USA}
  \postcode{19104}
}

\author{Oguzhan Demirag}
\email{odemirag17@posta.pau.edu.tr}
\affiliation{%
  \institution{Pamukkale University}
  \city{Denizli}
  \country{Turkey}
  \postcode{20010}
}

\author{Dilek Göksel Duru}
\email{dilek.goksel@tau.edu.tr}
\affiliation{%
  \institution{Turkish-German University}
  \state{Istanbul}
  \country{Turkey}
  \postcode{34000}
}

\author{Anup Das}
\email{anup.das@drexel.edu}
\affiliation{%
  \institution{Drexel University}
  \city{Philadelphia}
  \country{USA}
  \postcode{19104}
}


\renewcommand{\shortauthors}{Derebasi and Isik, et al.}

\begin{abstract}
    Guillain-Barre syndrome is a rare neurological condition in which the human immune system attacks the peripheral nervous system. A peripheral nervous system appears as a diffusively connected system of mathematical models of neuron models, and the system's period becomes shorter than the periods of each neural circuit. The stimuli in the conduction path that will address the myelin sheath that has lost its function are received by the axons and are conveying externally to the target organ, aiming to solve the problem of decreased nerve conduction. In the NEURON simulation environment, one can create a neuron model and define biophysical events that takes place within the system for study. In this environment, signal transmission between cells and dendrites are obtained graphically. The simulated potassium and sodium conductance are replicated adequately, and the electronic action potentials are quite comparable to those measured experimentally. In this work, we propose an analog and digital coupled neuron model comprising of individual excitatory and inhibitory neural circuit blocks for a low-cost and energy-efficient system. Compared to digital design, our analog design performs in lower frequency but gives a 32.3\% decreased energy efficiency. Thus, the resulting coupled analog hardware neuron model can be a proposed model for simulation of reduced nerve conduction. As a result, the analog coupled neuron, (even with its greater design complexity) serious contender for the future development of a wearable sensor device that could help with Guillain-Barre syndrome and other neurologic diseases.
\end{abstract}



\makeatletter
\renewcommand\@formatdoi[1]{\ignorespaces}
\makeatother
\maketitle

\section{Introduction}
Nervous system cells, which are the basic cells of the central nervous system, are called neurons. The human brain consists of about 86 billion neurons, whose primary role is to regulate all functions necessary for life using chemical and electrical signals \cite{azevedo2009equal}. Each neuron is connected densely to 50,000-250,000 other neurons, which are consisting of three types of neurons namely the afferent neuron, the efferent neuron and the interneuron. The electrical stimulation of the neurons occurs through the firing of the action potential, which is a rapid rise and fall of membrane potential. The action potential is generated when the membrane potential exceeds a given threshold value. The number of Na+ and Cl- ions is higher in the extracellular environment than in the intracellular environment, where the situation is the opposite for K+ ions.

\begin{figure}[h!]
    \centering
    \includegraphics[width=0.485\textwidth]{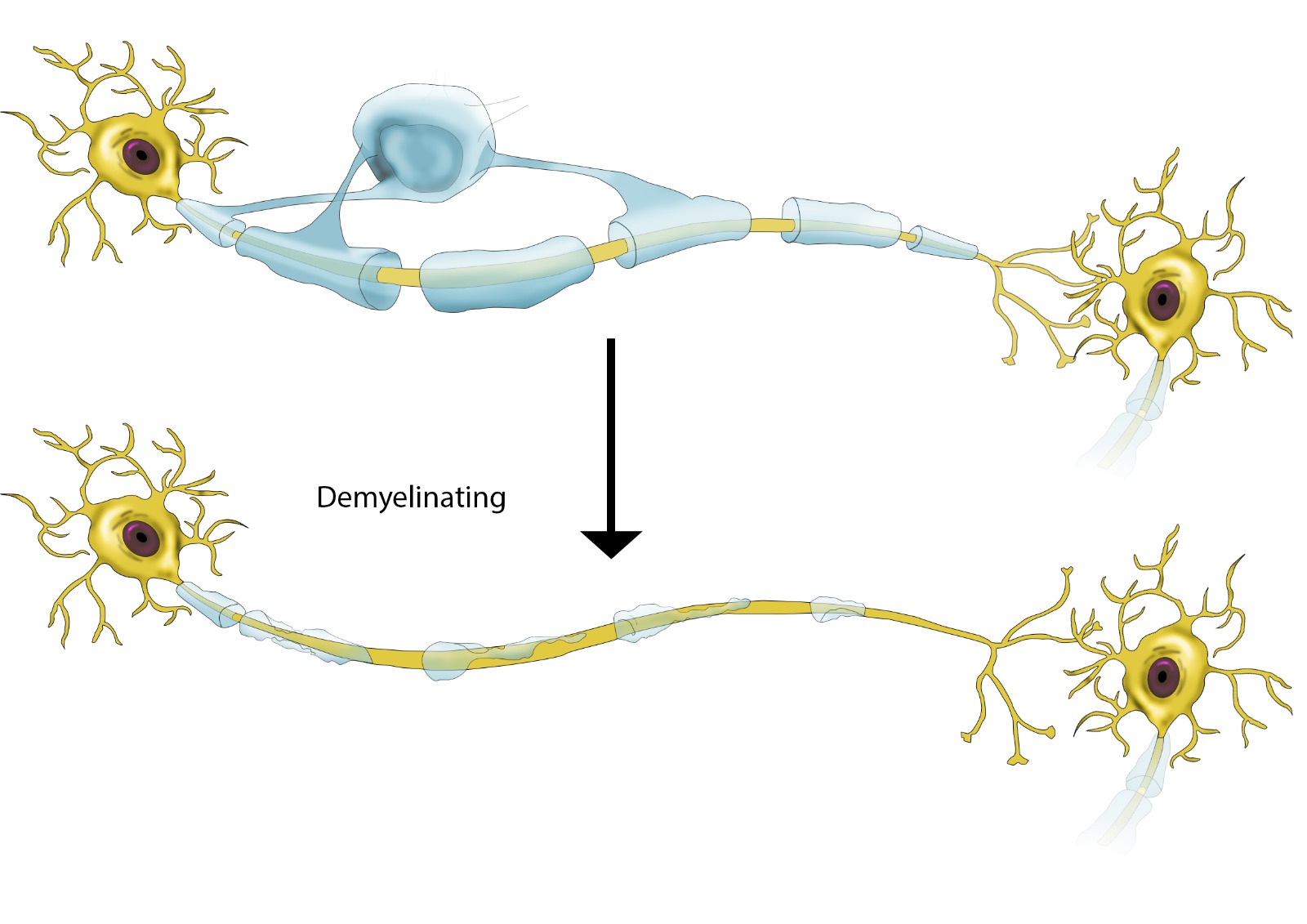}
    \caption{Demyelination}
    \label{fig_3}
\end{figure}

Myelin, which was coined by Rudolf Virchow in 1854, is the sheath that surrounds and protects the axons; 40\% of its mass consists of water, and the remainder is made up of 75\% lipid and 25\% protein. Along with its protective function, it also accelerates the transmission of and prevents the loss of impulses. It consists of the membrane of Schwann cells in the peripheral nervous system and the membrane of oligodendrocyte cells in the central nervous system. In cases, where myelin is damaged, a demyelinating disease is observed \nocite {miller2017animal}[27, \autoref{fig_3}]. Depending on the severity of demyelination, conduction may slow down or stop completely \cite{ironside1997oppenheimer,halliday1977pathophysiology}. Guillain-Barre Syndrome is an autoimmune neurological condition that causes temporary paralysis by damaging the peripheral nervous system. Muscle weakening leads to issues such as numbness and lack of reflexes. The immune system's antibodies assault the myelin coating in the nerve cell, causing it to rupture. It causes nerve conduction to slow or cease. 

The rest of the paper is organized as follows: \textbf{Section II}
presents the 3D Modelling of the neuron structure, the majority of neuron design choices, and the execution of each part of the neuron. \textbf{Section III} introduces Neuron Models. Hardware Implementation and its experimental results and analysis are reported in \textbf{Section IV}. \textbf{Section V} concludes the contents of this paper and \textbf{Section VI} give future aspects of this paper.

\section{3D Modelling}
The fact that neuron structures can be modeled mathematically has paved the way for simulating these models in a computer environment. The NEURON simulation program is at the forefront of modeling neuron models to the simulation environment. As of February 2021, more than 2400 scientific articles and book studies have been published using the NEURON simulation program \cite{Neuron1}. In its most general definition, NEURON is a simulation environment for developing and applying models of neurons and neuronal networks \cite{carnevale2006neuron}. It is suitable for describing and solving problems involving cable properties, near-membrane extracellular potential, and complex cell membrane properties that play an important role in describing neuron morphology. In addition, biophysical events, such as determining the intracellular and extracellular potassium, sodium, and calcium concentrations of the generated neurons, to determine the cell membrane potential can be defined and signal outputs can be obtained \cite{carnevale2006neuron}. In terms of programming, as well as the \emph{High Order Calculator} (HOC) interpreter, being programmable in the Python environment brings many conveniences and advantages. Python has powerful data structures, such as lists and dictionaries, built in to the language, and it has a large standard library, extensive functionality for data processing, database access, network programming, etc. \cite{hines2009neuron}.

\begin{figure}[h!]
    \centering
    \includegraphics[scale=0.5]{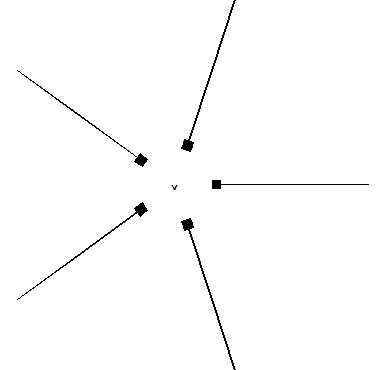}
    \caption{5 cells around 50 micron radius}
    \label{fig_5}
\end{figure}

An example of neuron morphology can be created and visualized in the NEURON program. After the biophysical properties of this model are entered, the output of the signals formed in the cells can be observed. An example neuron morphology with five cells is represented in \autoref{fig_5}.

In \autoref{fig_5}, there are ball and stick cells arranged in a ring. While performing the biophysical modeling of the cell, parameters, such as the diameter and length of the soma cell and the cell membrane element permeability, are defined. Finally, an axon from cell "n" connects to a synapse in the middle of the dendrite in cell "n+1". For this model, the specific dynamics of axons do not need to be explicitly modeled. When the soma fires an action potential, the spike is assumed to travel along the axon, and with some delay, induces a synaptic event in the dendrite of the target cell \cite{davison2009trends}. 

\begin{figure}[h!]
    \centering
    \includegraphics[width=0.485\textwidth]{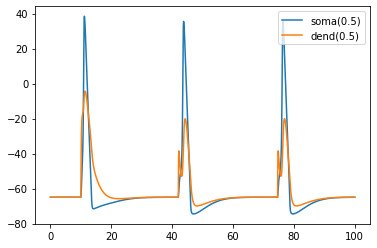}
    \caption{Signal transmission for a cell}
    \label{fig_6}
\end{figure}

\begin{figure}[h!]
    \centering
    \includegraphics[width=0.485\textwidth]{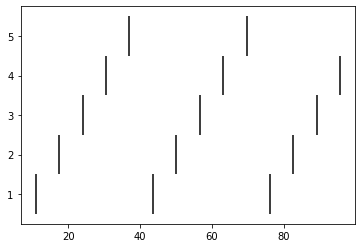}
    \caption{Signal sequence graph for all 5 cells}
    \label{fig_7}
\end{figure}

Multiple spikes of the signal in the cell in \autoref{fig_6} indicate that the spikes are transmitted throughout the network. The graph showing that all soma cells are fired in the sequence is illustrated in \autoref{fig_7}.

\begin{figure}[h!]
    \centering
    \includegraphics[width=0.3\textwidth]{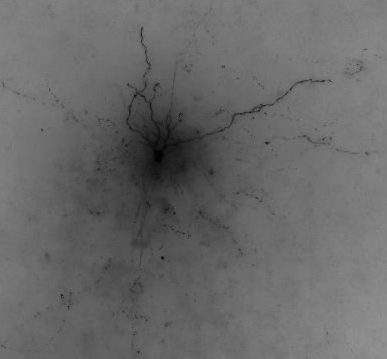}
    \caption{Real image of Gad2-IRES-Cre neuron from layer 5 from the Allen Mouse Brain Atlas}
    \label{fig_8}
\end{figure}

\begin{figure}[h!]
    \centering
    \includegraphics[width=0.485\textwidth]{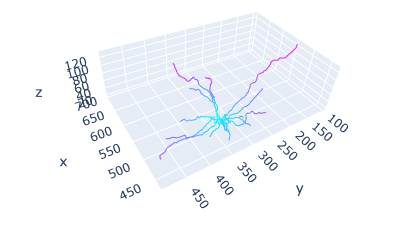}
    \caption{Gad2-IRES-Cre neuron from layer 5 in NEURON}
    \label{fig_9}
\end{figure}

In the NEURON program, the user can create his own cell morphology and topology, as well as imported pre-built models. In a study at the Allen Institute, a Gad2-IRES-Cre neuron from layer 5 of the mouse primary visual cortex was modeled with the NEURON program within the Allen Database of Cell Types. The model was based on a traced morphology after filling the cell with biocytin and optimized using experimental electrophysiology data recorded from the same cell. The electrophysiology data was collected in a highly standardized way to facilitate comparison across all cells in the database. The model was optimized by a genetic algorithm that adjusted the densities of conductance placed at the soma to match experimentally-measured features of action potential firing \cite{Atlas2004institute}. [10, \autoref{fig_8}] and \autoref{fig_9} show the real image of the neuron and its model created in the NEURON environment.

In the image in \autoref{fig_9}, the tracts of the neuron are divided into different segments (soma cell, dendrite, and axon). Then, using the positioning function, the color change was assigned according to the distance of the appendage in each section to the soma cell. Accordingly, the closest position to the soma cell is represented by light blue, and the farthest position is represented by dark purple \autoref{fig_9}. The exact location information is hidden inside the graph. This makes it easy to set values for all states at a particular location.

\begin{figure}[h!]
    \centering
    \includegraphics[width=0.485\textwidth]{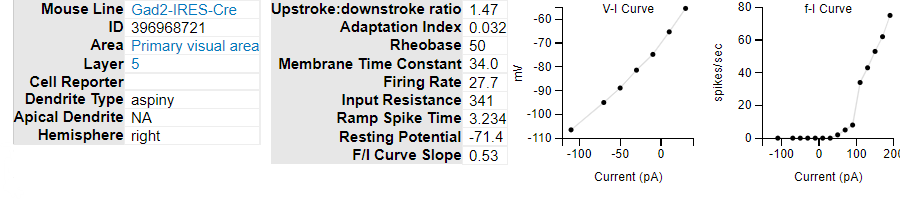}
    \caption{Electrophysiology summary}
    \label{fig_10}
\end{figure}

\begin{figure}[h!]
    \centering
    \includegraphics[width=0.485\textwidth]{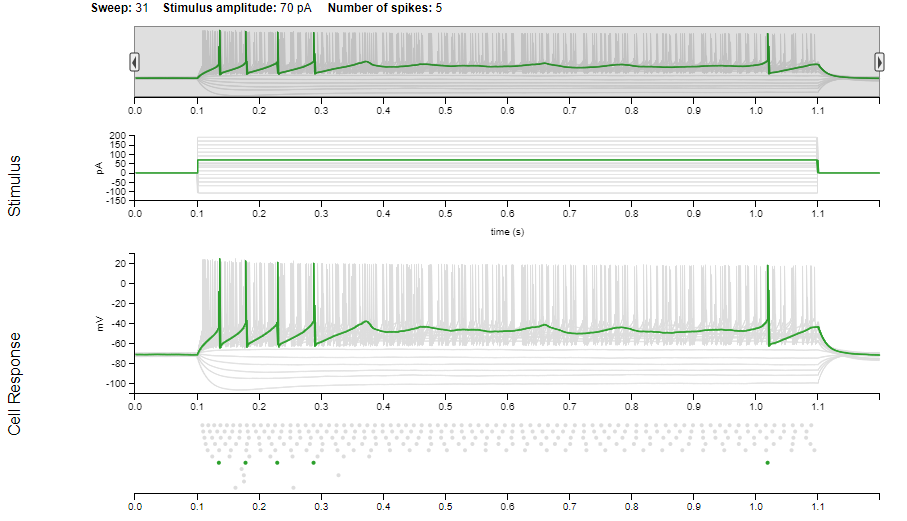}
    \caption{The output for 70pA stimulus amplitude}
    \label{fig_11}
\end{figure}

\begin{figure}[h!]
    \centering
    \includegraphics[width=0.485\textwidth]{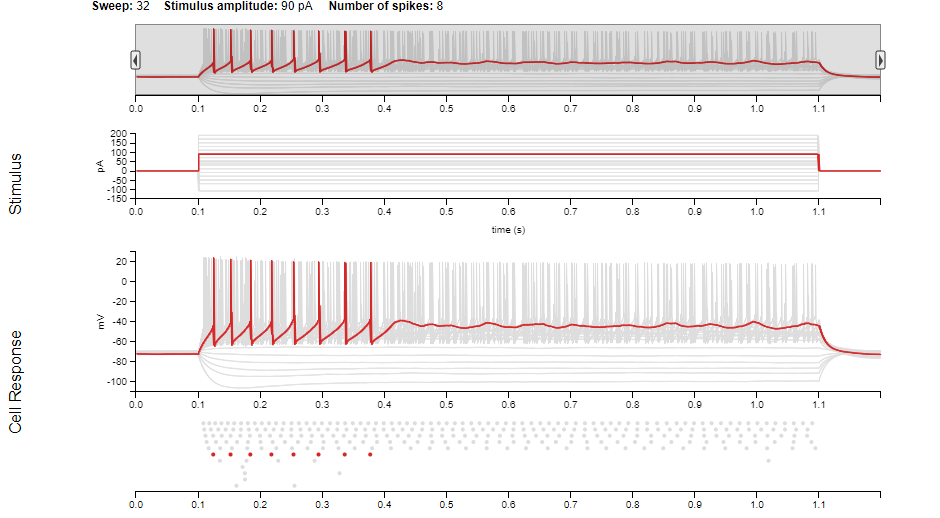}
    \caption{The output for 90pA stimulus amplitude}
    \label{fig_12}
\end{figure}

The simulation results obtained in the study are presented in \nocite {Atlas2004institute} [10, \autoref{fig_10}], [10, \autoref{fig_11}], and [10, \autoref{fig_12}]. The summary of electrophysiology is given in \autoref{fig_10}, where the values of neuron-specific morphological features are included. The first of the graphics in the image is the potential difference in the neuron depending on the pulse amplitude. The second graphic shows the spike flight time that changes depending on the current amplitude.

It is aimed to observe the behavior of the neuron cell applied with different pulse amplitudes. In this direction, it is observed that a total of 5 spikes occur when a pulse amplitude of 70pA is applied in \autoref{fig_11}, and a total of 8 spikes are formed when a pulse amplitude of 90pA is applied in \autoref{fig_12}. As the pulse amplitude is increased, the observed spike also increases.

\section{Neuron Models}
\subsection{Hodgkin-Huxley Neuron Model}
\begin{figure}[h!]
    \centering
    \includegraphics[width=0.35\textwidth]{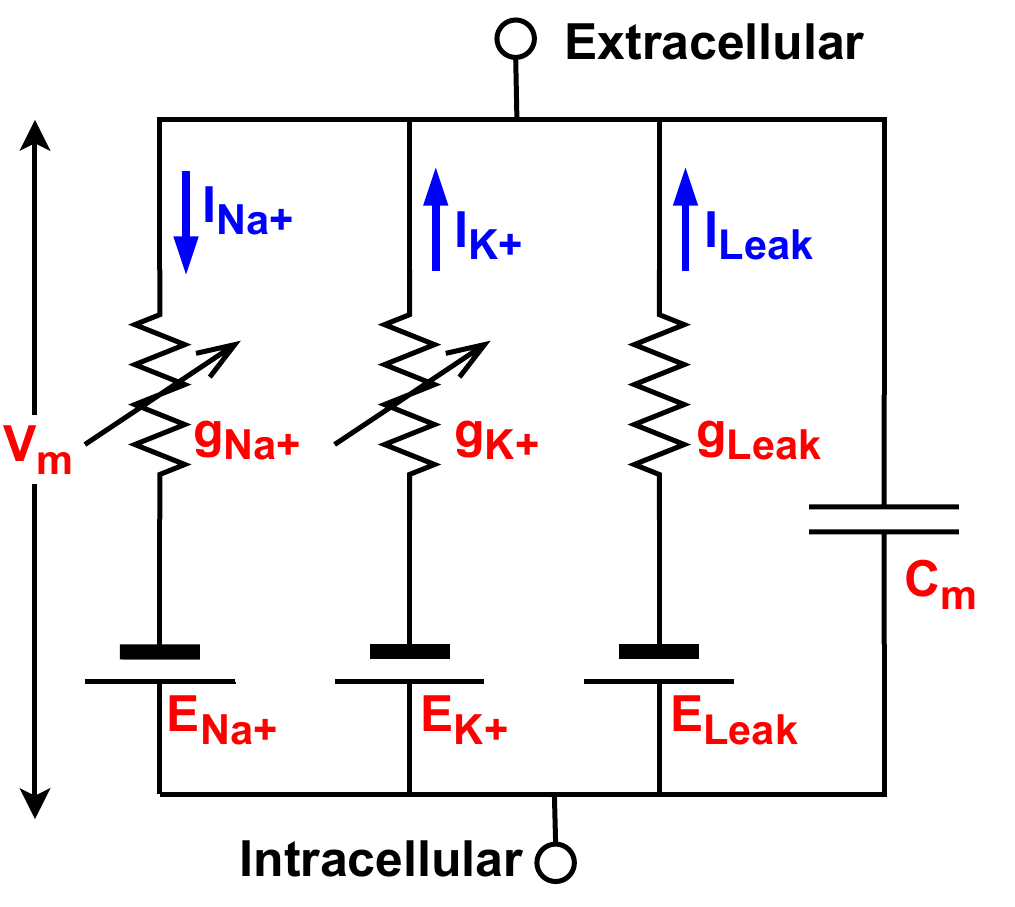}
    \caption{Hodgkin-Huxley (HH) Model Equivalent Circuit}
    \label{fig_4}
\end{figure}

The \emph{Hodgkin-Huxley} (HH) neuron model is the study that explains the ionic mechanism that provides the generation and transmission of action potentials on the squid giant axon \cite{hodgkin1952quantitative}.
The HH Model forms the basis of many existing neuron models. This model is based on the assumption that the ion channels in the membrane work independently of each other and that each channel passes a certain ion. Leakage channels with no specific selectivity are assumed to carry different ion types. The equivalent schematic of the HH Model is shown in \autoref{fig_4}.

For channels, the ion permeability of the membrane is found according to Ohm's law using the following formulas:

\begin{equation} 
    G_{Na}=\frac{I_{Na}}{V_m-V_{Na}}
\end{equation}
\begin{equation} 
    G_K=\frac{I_K}{V_m-V_K}
\end{equation}
\begin{equation} 
    G_L=\frac{I_L}{V_m-V_L}
\end{equation}

The sum of the membrane current is represented by Kirchhoff's current law as follows:
\begin{equation} 
    I_m=C_m\frac{dV_m}{dt}+(V_m{-}V_N{}_a)G_N{}_a{+}(V_m{-}V_k)G_K+(V_m{-}V_L)G_L
\end{equation}\newline

\subsection{Izhikevich Neuron Model}

This model has emerged by reducing the HH Neuron model, which includes 4 varying differential equations, and is valuable in terms of its computational ease and rich neuron behavior \cite{izhikevich2003simple}. Differential equations defining the Izhikevich neuron model are given in \eqref{eqn:5}, \eqref{eqn:6} and \eqref{eqn:7}:

\begin{equation}
    v = 0.04^{2}+5v+140-u+I
    \label{eqn:5}
\end{equation}

\begin{equation}
    u = a(bv-u)
    \label{eqn:6}
\end{equation}

\begin{equation}
    v => 30mV \implies
    \binom{v \leftarrow c}{u \leftarrow u+d}
    \label{eqn:7}
\end{equation}

Different neuron dynamics can be exhibited when fixed parameter values are properly adjusted. Therefore, it is a suitable model for the study of neuron behavior.\\

\subsection{FitzHugh-Nagumo Neuron Model}
This neuron model has simplified the HH neuron model actually, where the voltage gate variables' dependency on each other is found. As a result of these studies, he observed that the voltage gate variables were related to each other. 
HH made changes to the "Van Der Pol Oscillator Model in order to simplify the Neuron Model. The \emph{FitzHugh-Namugo} (FHN) neuron model  formed as a result of the studies is explained with the Equations \eqref{eqn:8} and \eqref{eqn:9} below \cite{fitzhugh1961impulses}:

\begin{equation}
    \frac{dv}{dt} = c(v-u+I-\frac{v^{3}}{3})
    \label{eqn:8}
\end{equation}

\begin{equation}
    \frac{du}{dt} = \frac{(v-bu+a)}{c}
    \label{eqn:9}
\end{equation}

FHN Neuron Model performs only spike behavior among neuron dynamics. The frequency change that occurs with the change of the “I” value in the HH Neuron Model is absent in this model.

\subsection{Hindmarsh-Rose Neuron Model}
The FHN model is a simple version of the HH model and cannot exhibit all neuron dynamics. The \emph{Hindmarsh-Rose} (HR) neuron model was obtained by adding another state variable to the bivariate equation of the FHN Neuron Model. It is a simple model but can exhibit most of the neuron dynamics. It is defined by the following equations \eqref{eqn:10}, \eqref{eqn:11} and \eqref{eqn:12}:

\begin{equation}
    v'=u-v^{3}+bx^{2}+I-w
    \label{eqn:10}
\end{equation}

\begin{equation}
    u= 1-5v^{2}-u
    \label{eqn:11}
\end{equation}

\begin{equation}
    w=\mu(s(v-v_{rest})-w)
    \label{eqn:12}
\end{equation}

\subsection{Morris-Lecar Neuron Model}
\emph{Morris-Lecar} (M-L) neuron model is a model using calcium, potassium, and ohmic load currents and expresses muscle fibers mathematically. It is expressed by the differential equations \eqref{eqn:13} and \eqref{eqn:14}:

\begin{equation}
    C\frac{dV}{dt} = I-g_{L}(V-V_{L})-g_{Ca}M_{ss}(V-V_{Ca})-g_{K}N(V-V_{K})
    \label{eqn:13}
\end{equation}

\begin{equation}
    \frac{dN}{dt}=\frac{N_{SS}-N}{\tau_{N}}
    \label{eqn:14}
\end{equation}
\\
\subsection{Coupled Neuron Models}
Neurons communicate and coordinate through synapses. This coupling between neurons creates scenarios where rapid action, (such as a working reflex). Coupling neurons connect with each other using electrical and chemical pathways. Since the electrical coupling is fast in nature, electrical conduction is used to replicate this effect such as reflex. A single neuromorphic hardware can implement only a limited number of neurons and synapses \cite{icsik2022design}. \textbf{For this work, M-L, FHN, and HR Neuron Models (which are described below) were used to study the behavior of neuron coupling.}
\begin{itemize}
    \item The M-L model, which was put forward as the mathematical modeling of muscle fibers, was arranged as in equation \eqref{eqn:15}, \eqref{eqn:16}, \eqref{eqn:17}, \eqref{eqn:18}, \eqref{eqn:19} in a research paper to coupling. In that study, synaptic (chemical) coupling between 2 neurons whose neuron model was implemented with hardware was created and examined \cite{xue2008inhibitory}.
    
\begin{equation}
\begin{split}
    C\frac{dV_{1}}{dt} = -g_{ca}M_{\infty}(V_{1})-(V_{1}-V_{ca})-g_{K}M_{\infty}(V_{1})-(V_{1}-V_{ca})\\
    -g_{L}M_{\infty}(V_{1})-(V_{1}-V_{L})+1 
    \label{eqn:15}
\end{split}
\end{equation}

\begin{equation}
\frac{dN_{1}}{dt}\lambda(-N_{1}+(G(V_{1}))
\label{eqn:16}
\end{equation}

\begin{equation}
\tau\frac{dZ}{dt}=\alpha F(V_{1})-Z
\label{eqn:17}
\end{equation}

\begin{equation}
\begin{split}
    C\frac{dV_{2}}{dt} = -g_{ca}M_{\infty}(V_{2})-(V_{2}-V_{ca})-g_{K}M_{\infty}(V_{2})-(V_{2}-V_{ca})\\
    -g_{L}M_{\infty}(V_{2})-(V_{2}-V_{L})+1-Z(V_{2}-\gamma)
\label{eqn:18}
\end{split}
\end{equation}

\begin{equation}
\frac{dN_{2}}{dt}\lambda(-N_{2}+(G(V_{2}))
\label{eqn:19}
\end{equation}

    \item The mathematical representation of the FHN model was modified to study the behavior of synchronized neurons. Modified mathematical are represented in equations \eqref{eqn:20}, \eqref{eqn:21}, \eqref{eqn:22}, \eqref{eqn:23}.   
    
    \begin{equation}
    \frac{dV_{1}}{d{\tau}} =\begin{bmatrix}V_{1}-\left(\frac{V_{1}^3}{3}\right)\end{bmatrix}-{W_{1}}
    \label{eqn:20}
\end{equation}

\begin{equation}
    \frac{dW_{1}}{d{\tau}}=	\epsilon_{1}\begin{bmatrix}g(V_{1})-W_{1}-\eta_{1}\end{bmatrix}
    \label{eqn:21}
\end{equation}

\begin{equation}
    \frac{dV_{2}}{d{\tau}} =\begin{bmatrix}V_{2}-\left(\frac{V_{2}^3}{3}\right)\end{bmatrix}-{W_{2}}+DV_{1}
    \label{eqn:22}
\end{equation}

\begin{equation}
    \frac{dW_{2}}{d{\tau}}=
    \epsilon_{2}\begin{bmatrix}g(V_{2})-W_{2}-\eta_{2}\end{bmatrix}
    \label{eqn:23}
\end{equation}

    \item One of the reported models suitable for studying the behavior of synchronized neurons is the Hindmarsh-Rose neuron model. In the coupled HR neuron model, the interaction between "i" and "j" neurons are expressed using the following equations \eqref{eqn:24}, \eqref{eqn:25}, \eqref{eqn:26}.

    \begin{equation}
    v'=u{-}v^{3}-bv{+}I{-}w{-}g_{s}\sigma(v_{i})\displaystyle\sum\limits_{j{=}1}^{N} {C_{ij}} \gamma(V_{i}{,}V_{j})
    \label{eqn:24}
\end{equation}
\begin{equation}
    u'= 1-5v^{2}-u
    \label{eqn:25}
\end{equation}

\begin{equation}
    w'=\mu(s(v-v_{rest})-w)
    \label{eqn:26}
\end{equation}
\end{itemize}

\section{Hardware Implementation}
\textbf{Proposed circuit model is aimed to represent the stimuli that cannot be transmitted or incompletely transmitted due to the damaged sheath and to ensure that they are transmitted to the target neurons intact. Since the model is designed to be implemented through assistive devices, the advantage of implementation has been taken into account. The coupled \emph{Hardware Neuron} (HN) model is preferred, which is suitable for generating an artificial axon. The HN model provides an advantage in terms of implementation. Protection of stimulus is ensured by acting as the myelinated axon body}. Our circuit's fundamental concept can be applied to many different types of membranes and can generate a wide range of electrical responses. There are a number of benefits to studying HN models rather than merely mathematical formulas. \textbf{Real-time processing is possible with an HN model and brings for instance energy efficiency, more accurate and multi-tasking}. As a result, when several HN models are coupled to each other, it would be suitable for researching large-scale neural network dynamics. Each parameter of a model has a clear physical interpretation, and in certain circumstances, it can be directly correlated with electrophysiological values of membranes. Circuit tests and mathematical analysis to gain a better understanding of the dynamics. These dynamics separate parts of the design process such that they can be independently optimized for different metrics such as performance, power, cost, and reliability \cite{huynh2022implementing}.

\subsection{Analog Design}
\begin{figure}[h!]
    \centering
    \includegraphics[width=0.485\textwidth]{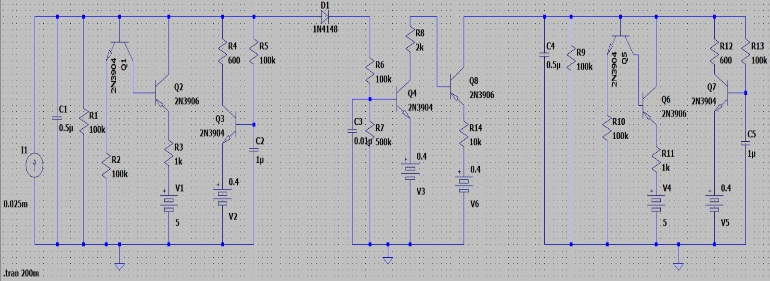}
    \caption{Excitatory Circuit}
    \label{Excitatory Circuit}
\end{figure}

\begin{figure}[h!]
    \centering
    \includegraphics[width=0.485\textwidth]{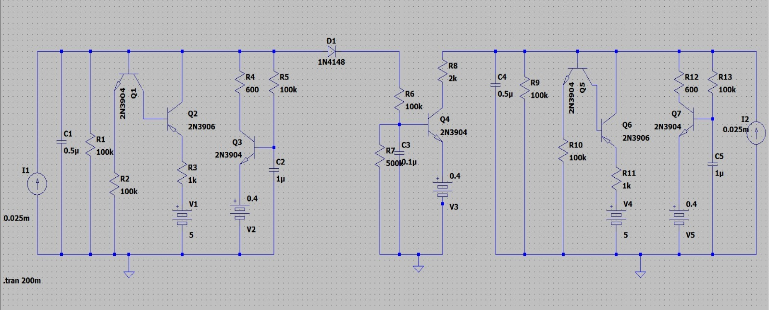}
    \caption{Inhibitory Circuit}
    \label{Inhibitory Circuit}
\end{figure}

Two neuromorphic circuits are compared in this study, and these circuits are performing for synchronization HN models. An acceleration phenomenon appears in a diffusively connected system of mathematical models of the axon, with the system's period becoming shorter than the periods of each isolated circuit \cite{maeda2005synchronization}. Axons self-organize in such a way that they accelerate through axon and body cell coupling while consuming the least amount of energy. Our target is used for inhibitory synapse and excitatory synapse in analog design. 
\textbf{In our circuit design, an excitatory synapse which is shown in Figure \ref{Excitatory Circuit} is formed by modifying the depolarizing process channel to be dependent on presynaptic voltage (and altering the reversal potential). The top trace represents a presynaptic cell. The diode in the synapse symbolizes the presynaptic side's isolation as a result of transmitter release. An inhibitory synapse is shown in Figure \ref{Inhibitory Circuit} is formed by modifying the repolarizing process channel to be dependent on presynaptic voltage. Shunting and hyperpolarization are observed at this inhibitory synapse. By connecting the two cells using a diode and resistors, an electrotonic link is formed, with a stronger connection from left to right. The traces demonstrate a highly complicated relationship between the two cells.}

The rectified electrotonic synapse can be rectified in either direction or not at all. \emph{Negative-Positive-Negative} (NPN) and \emph{Positive-Negative-Positive} (PNP) transistors occupy an important place. Our selection are 2N3904 type of transistor which is a very common NPN transistor that can take 200 mA (absolute maximum) and frequencies as high as 100 MHz when used as an amplifier and 2N3906 type of transistor which is a typical PNP bipolar junction transistor used for general-purpose, low-power amplification or switching. It is intended for modest electric current and power, medium voltage, and relatively high speeds. The dynamics of our model are based on hardware experimentation and mathematical studies. A generic type of neural circuit have been proposed in this study:

Typical dynamics in the HN model of Maeda \cite{tsai2018circuit}. The parameters are C=0.5 mF, \(C_n\)=1.0 mF, \(R_L=100 k\si{\ohm}\), \(R_1=200 k\si{\ohm}\), \(R_2=2 k\si{\ohm}\), \(R_3=100 k\si{\ohm}\), \(E_Na\)= 5 V, \(E_K\)= −0.4 V, \(E_L\)= 0 V, \(T_1\)=\(T_3\): NPN-type 2N3904, T2: PNP-type 2N3906, and \(I_ext\)=0.025 mA. The ordinate is the membrane potential V. In the literature, one may conduct circuit experiments as well as quantitative analysis to gain a better knowledge of the dynamics. Several HN model circuits have been built with motives comparable to recent models \cite{gulrajani1977modelling}\cite{sekine1999study}.

Excitation and inhibition have been widely investigated in terms of proper neuronal function and stability, and disruption of excitation and inhibition has been linked to a variety of neurological diseases. When two neuron activation function circuits are manually linked into the excitation and inhibition module circuit, hardware experiments for the adapting synapse-based neuron model can be performed. We built excitatory-inhibitory circuits by adding excitatory nodes and edges to simulate synaptic diversity. This diversity attaches additional connection circuit edges in combination with the addition of inhibitory + excitatory nodes and expands the encoding capacity of our network. It is shown in \ref{}

\begin{figure}[h!]
    \centering
    \includegraphics[width=0.460\textwidth]{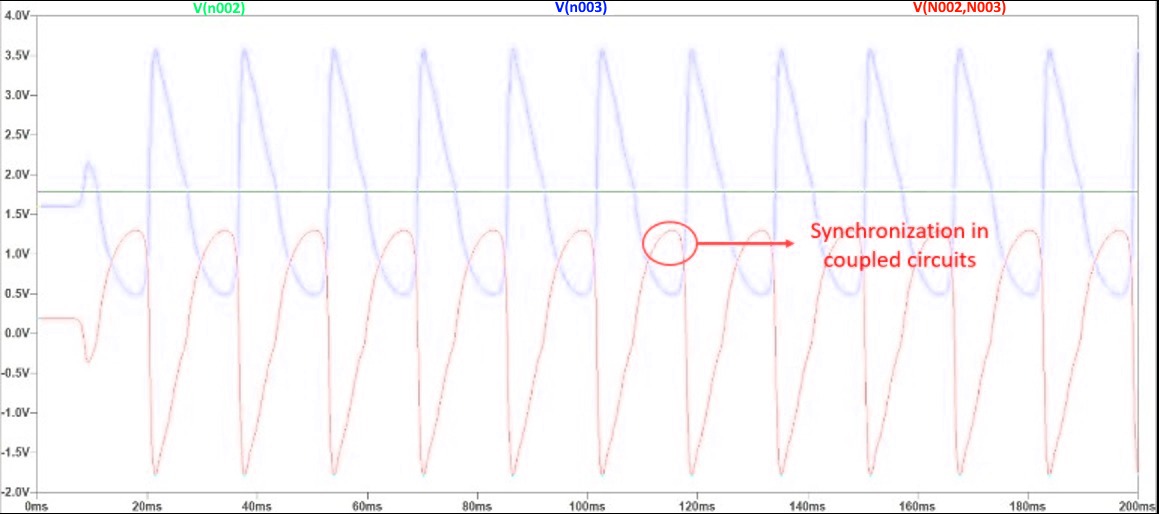}
    \caption{Simulation Results of Proposed Circuit}
    \label{fig_2_FPGA_block}
\end{figure}
\begin{figure}[h!]
    \centering
    \includegraphics[width=0.480\textwidth]{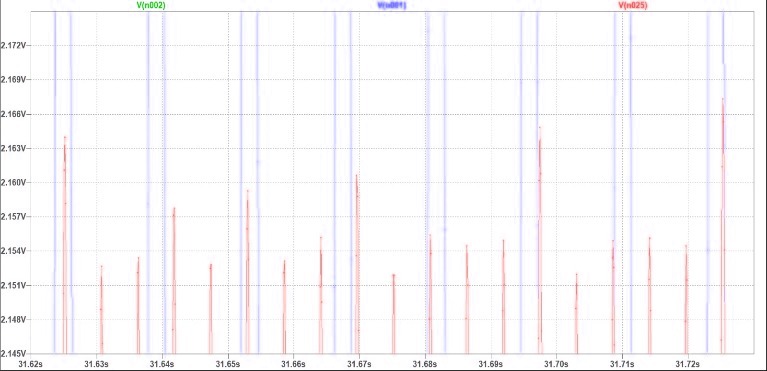}
    \caption{Simulation Results of Proposed Circuit}
    \label{Simulation Results of Proposed Circuit}
\end{figure}

The output impedance of the inhibitory circuit, the \(R_c\) resistor, and a 1uF capacitor are added to the input impedance of the excitatory circuit. The expectation here was, from the roles of each parameter, that the former would be associated with the modification of firing patterns and the latter with the generation of the potential voltage. In this way, it is aimed that the circuits work in accordance with the coupled HN model. The secondary circuit resembled the synaptic voltage value that appeared at the conclusion of the first circuit at 2.10 Watt values which are shown in Figure \ref{Simulation Results of Proposed Circuit}, and it was demonstrated that the circuits could operate in a suitable manner.

This is significant because of the measurement, and hence the definition of excitatory and inhibitory, may be related to its function. Excitatory and inhibitory roles have been hypothesized in a variety of ways. It is stated that the temporally tightly matched E and I circuits described in sensory cortices can only be measured by sensory-evoked synaptic responses and may be essential for efficient sensory information coding, whereas the balance of E and I at the level of single neurons and circuits, in general, may safeguard circuit stability in previous works \cite{o2017beyond}.  
It demonstrated two ways for building a varied excitatory-inhibitory neural network, starting with network topology and then introducing different types of inhibitory interneurons and circuit variability to the simulated network. They discovered that inhibitory interneurons improve encoding capacity by shielding the network from extremely brief activation periods when network wiring complexity is high. Furthermore, various kinds of interneurons have varying impacts on encoding capacity and reliability \cite{deneve2016efficient}. Our proposed circuit measured area with die shrink methodology which is the act of an identical circuit, using a more advanced manufacturing process. 

\subsection{Digital Design}
To demonstrate excellent design principles, the hardware neural system was implemented on Xilinx Kria KV260 FPGA. The mentioned system produces a hardware platform comprising a high-performance XCK26-SFVC784-2LV FPGA chip. The number of hardware neurons specified in this study is realized in a single core of the KV260 development board. This board represented the latest product of \emph{System on Module} (SoM). Because of its reconfigurability, developers may create apps with little or no external memory accesses, which not only reduces the total power consumption of the program but also improves responsiveness with lower end-to-end latency \cite{vision2021kria}. The use of artificial neural networks is critical in order to analyze massive amounts of data collected from millions of electronic devices all around the world. This sector is under constant improvement since it is critical to increase speed and performance in order to match the defined restrictions and objectives.

\begin{figure}[h!]
    \centering
    \includegraphics[width=0.485\textwidth]{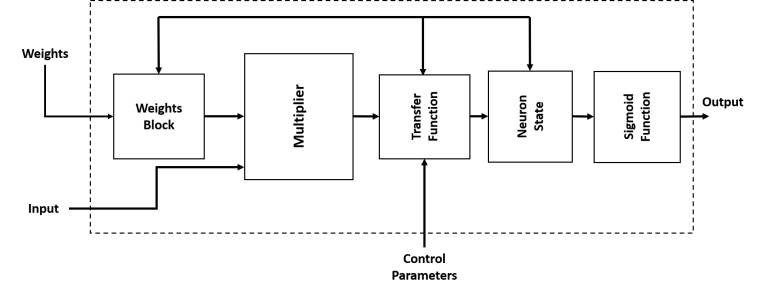}
    \caption{Structure of Hardware Neuron Model}
    \label{fig_2_FPGA_block}
\end{figure}

Capacitor weights are small; however, they have leakage issues and are substantial to prevent weight decay. Capacitor weights with refreshment can help with leakage issues, but they need off-chip memory. Digital multipliers use 130 nm technology and have a larger area than analog multipliers. It necessitates a huge number of multiplications, which can be multiplexed onto a small number of physical-digital multipliers. The total of the results of multiplications yields a weighted input signal. The transfer function is performing on/off the chip, and it controls and evaluates the weighted sum of the neural inputs.  The sigma activation block, which is constantly on the neuro-chip, analyzes the neuron state block of the transfer function. The fastest way to implement the sigmoid function was using non-volatile, read-only memories. The sigmoid function is one of the most often utilized activation functions since it fluctuates between 0 and 1, as are probabilities, which are the outputs of neurons. An activation function is applied to a weighted sum of inputs, and the outcome is used as an input to the following layer. The output of this unit will be a nonlinear function of the weighted input sum since the sigmoid is a nonlinear function. A neuron that uses a sigmoid function as an activation function is called a sigmoid unit. Digital design results are shown in Table \ref{table:table_1}. 

\begin{table}[h!]
\centering
\caption{Utilization Summary}
\label{table:table_1}
\begin{tabular}{|c|c|c|c|c|}
\hline
Logic Utilization & \begin{tabular}[c]{@{}c@{}}Used\\ \end{tabular} & \begin{tabular}[c]{@{}c@{}}Available\\ \end{tabular} & \begin{tabular}[c]{@{}c@{}}Device\\ \end{tabular} \\ \hline

LUTs & 2067 & 11720 & XCK26-SFVC784-2LV\\ \hline
Registers& 112 & 2344240 & XCK26-SFVC784-2LV\\ \hline
DSPs& 5 & 1248 & XCK26-SFVC784-2LV\\ \hline
GCLKs & 1 & 352 & XCK26-SFVC784-2LV\\ \hline
Power & 2.780W & - & XCK26-SFVC784-2LV\\ \hline

\end{tabular}
\end{table}

\subsection{Overall Performance}
We have demonstrated that analog model consumes lower power estimation is that computed to the digital model. The literature \cite{vittoz1994low} clearly states that the minimal power consumption in analog circuitry is entirely dictated by the acquired precision and operating frequency. It should be noted that the circuitry used to program the floating gate charge consumes a significant amount of chip space (on-chip charge-pumps or high-voltage selection transistors).
The accuracy of an analog circuit is typically described in terms of \emph{Signal-to-Noise-Ratio} (SNR), which is the ratio of signal power to noise power. In real circuits, the SNR inaccuracy is the ratio of signal to total of all noise, mismatch, drifts, and so on. We have presented only power consumption and synchronization in our work.


\begin{table*}[h!]
\centering
\caption{Comparison of the proposed implementations with the related works for \underline{small moderate scale} models}
\label{table:table_2}
\begin{tabular}{|l|l|l|l|l|l|}
\hline
                  &Work  &Technology  &Design Methodology  &Power  &Operation Frequency  \\ \cline{2-6}
&Grassia et al.~\cite{grassia2016digital}  &\SI{65}{\nano\meter}  &Digital design  &$\approx$ \SI{4.560}{\watt}&100 MHz  \\\cline{2-6}
                  &Joubert et al.~\cite{joubert2012hardware}  &\SI{65}{\nano\meter}  &Analog design  &\SI{78.160}{\nano\watt} &256 MHz  \\\cline{2-6} 
                  &Cosp et al.~\cite{cosp2014realistic}  &\SI{0.35}{\micro\meter}  &Analog design  &\SI{40.95}{\nano\watt} &100 MHz  \\\cline{2-6}
                  &Khanday et al.~\cite{khanday2018low}  &\SI{130}{\nano\meter} &Analog design  &\SI{22.8}{\micro\watt}&8.98 MHz \\\cline{2-6}
                  
                  & \textbf{Our Work }&\SI{16}{\nano\meter}  &Digital design  &\SI{2.780}{\watt}  &100 MHz    \\\cline{3-6} &&\SI{2}{\micro\meter}  &Analog design  &\SI{2.100}{\watt}  &1 MHz \\\hline
\end{tabular}
\end{table*}

Table \ref{table:table_2} shows the overall performance of various design approaches. All studies are evaluated on small moderate scale design instead of neuromorphic chips (e.g., SpiNNaker \cite{furber2014spinnaker} , TrueNorth \cite{merolla2014million} , Loihi \cite{davies2018loihi} , BrainScaleS-1 \cite{meier2015mixed}, NeuroGrid \cite{benjamin2014neurogrid} , DYNAP \cite{moradi2017scalable} , ODIN \cite{frenkel20180} ) . Thus, in spite of its limited application, it is possible to build coupled circuits for this disease in a low-power and low-price manner. Furthermore, for a variety of classification applications, an analog feed-forward neural network (rather than digital neural nets) is chosen for low power consumption. Analog neural networks consume less power. \textbf{As a result, our proposed circuit lowered 32,3\% of the power usage and used the lowest-price transistors in the market.}  One of the reasons for the acceleration is that the connected system's duty cycle is as low as possible. We built hardware membrane models that simulated the action potential of neurons in order to investigate synchronization in a diffusively coupled system. When two circuits with the same working principle (algorithm) but differing progress lengths were connected together, the resulting system gave rise to the acceleration HN model concept.
\section{Conclusions}

Contrary to our assumptions, we discovered that coupling circuits might minimize power usage. In other words, the Coupled Neuron Model efficiently drove the power usage described by the proposed neural circuit. The impulse produced by the neuron cell was successfully copied and the output was taken from the analog circuit, which gave the most efficient result. This output can be correctly transmitted to the target organ with the application. The generated impulses are coupled and shown in the simulation environment. Accompanied by coupled HN model, it is modeled to increase the functionality of the patient by providing the transmission of balance and coordination reflexes that require speed transmission.

\section{FUTURE ASPECTS}
Our experience is this proposed HN model based on both analog and digital designs. These designs are getting mixed-signal design options to show that simulation works. This research opens the possibility of creating a wearable sensor device that could help on Guillain-Barre syndrome and other neurological diseases. We would like to expand our model to a wearable sensor level. It has been observed, that a response can occur in body cells that are not innervated as a result of Guillain-Barre syndrome. In addition, it is planned to test the proposed circuit to transfer from the sensory area to the brain together with the reverse transformation methodology, and it is planned as future studies that the sensory information in the receptor cells can be transferred to the central nervous system. Future studies will cover that our model will be produced and tested for progress in the lab environment.

\section{ACKNOWLEDGEMENT}
We thank Lakshmi Mirtinti for her technical assistance and for her valuable contributions.

\bibliographystyle{IEEEtranSN}
\bibliography{sample-base}

\begin{thebibliography}{34}
\providecommand{\natexlab}[1]{#1}
\providecommand{\url}[1]{#1}
\csname url@samestyle\endcsname
\providecommand{\newblock}{\relax}
\providecommand{\bibinfo}[2]{#2}
\providecommand{\BIBentrySTDinterwordspacing}{\spaceskip=0pt\relax}
\providecommand{\BIBentryALTinterwordstretchfactor}{4}
\providecommand{\BIBentryALTinterwordspacing}{\spaceskip=\fontdimen2\font plus
\BIBentryALTinterwordstretchfactor\fontdimen3\font minus
  \fontdimen4\font\relax}
\providecommand{\BIBforeignlanguage}[2]{{%
\expandafter\ifx\csname l@#1\endcsname\relax
\typeout{** WARNING: IEEEtranSN.bst: No hyphenation pattern has been}%
\typeout{** loaded for the language `#1'. Using the pattern for}%
\typeout{** the default language instead.}%
\else
\language=\csname l@#1\endcsname
\fi
#2}}
\providecommand{\BIBdecl}{\relax}
\BIBdecl

\bibitem[Azevedo et~al.(2009)Azevedo, Carvalho, Grinberg, Farfel, Ferretti,
  Leite, Filho, Lent, and Herculano-Houzel]{azevedo2009equal}
F.~A. Azevedo, L.~R. Carvalho, L.~T. Grinberg, J.~M. Farfel, R.~E. Ferretti,
  R.~E. Leite, W.~J. Filho, R.~Lent, and S.~Herculano-Houzel, ``Equal numbers
  of neuronal and nonneuronal cells make the human brain an isometrically
  scaled-up primate brain,'' \emph{Journal of Comparative Neurology}, vol. 513,
  no.~5, pp. 532--541, 2009.

\bibitem[Benjamin et~al.(2014)Benjamin, Gao, McQuinn, Choudhary,
  Chandrasekaran, Bussat, Alvarez-Icaza, Arthur, Merolla, and
  Boahen]{benjamin2014neurogrid}
B.~V. Benjamin, P.~Gao, E.~McQuinn, S.~Choudhary, A.~R. Chandrasekaran, J.-M.
  Bussat, R.~Alvarez-Icaza, J.~V. Arthur, P.~A. Merolla, and K.~Boahen,
  ``Neurogrid: A mixed-analog-digital multichip system for large-scale neural
  simulations,'' \emph{Proceedings of the IEEE}, vol. 102, no.~5, pp. 699--716,
  2014.

\bibitem[Carnevale and Hines(2006)]{carnevale2006neuron}
N.~T. Carnevale and M.~L. Hines, \emph{The NEURON book}.\hskip 1em plus 0.5em
  minus 0.4em\relax Cambridge University Press, 2006.

\bibitem[Carnevale()]{Neuron1}
T.~Carnevale, ``The neuron bibliography.''

\bibitem[Cosp et~al.(2014)Cosp, Binczak, Madrenas, and
  Fern{\'a}ndez]{cosp2014realistic}
J.~Cosp, S.~Binczak, J.~Madrenas, and D.~Fern{\'a}ndez, ``Realistic model of
  compact vlsi fitzhugh--nagumo oscillators,'' \emph{International Journal of
  Electronics}, vol. 101, no.~2, pp. 220--230, 2014.

\bibitem[Davies et~al.(2018)Davies, Srinivasa, Lin, Chinya, Cao, Choday, Dimou,
  Joshi, Imam, Jain, et~al.]{davies2018loihi}
M.~Davies, N.~Srinivasa, T.-H. Lin, G.~Chinya, Y.~Cao, S.~H. Choday, G.~Dimou,
  P.~Joshi, N.~Imam, S.~Jain \emph{et~al.}, ``Loihi: A neuromorphic manycore
  processor with on-chip learning,'' \emph{Ieee Micro}, vol.~38, no.~1, pp.
  82--99, 2018.

\bibitem[Davison et~al.(2009)Davison, Hines, and Muller]{davison2009trends}
A.~P. Davison, M.~Hines, and E.~Muller, ``Trends in programming languages for
  neuroscience simulations,'' \emph{Frontiers in neuroscience}, vol.~3, p.~36,
  2009.

\bibitem[Den{\`e}ve and Machens(2016)]{deneve2016efficient}
S.~Den{\`e}ve and C.~K. Machens, ``Efficient codes and balanced networks,''
  \emph{Nature neuroscience}, vol.~19, no.~3, pp. 375--382, 2016.

\bibitem[FitzHugh(1961)]{fitzhugh1961impulses}
R.~FitzHugh, ``Impulses and physiological states in theoretical models of nerve
  membrane,'' \emph{Biophysical journal}, vol.~1, no.~6, pp. 445--466, 1961.

\bibitem[for Brain~Science(2004)()]{Atlas2004institute}
A.~I. for Brain~Science(2004), ``Allen mouse brain atlas [dataset].''

\bibitem[Frenkel et~al.(2018)Frenkel, Lefebvre, Legat, and Bol]{frenkel20180}
C.~Frenkel, M.~Lefebvre, J.-D. Legat, and D.~Bol, ``A 0.086-{$\mathrm{mm}^{2}$}
  12.7-pj/sop 64k-synapse 256-neuron online-learning digital spiking
  neuromorphic processor in 28-nm cmos,'' \emph{IEEE transactions on biomedical
  circuits and systems}, vol.~13, no.~1, pp. 145--158, 2018.

\bibitem[Furber et~al.(2014)Furber, Galluppi, Temple, and
  Plana]{furber2014spinnaker}
S.~B. Furber, F.~Galluppi, S.~Temple, and L.~A. Plana, ``The spinnaker
  project,'' \emph{Proceedings of the IEEE}, vol. 102, no.~5, pp. 652--665,
  2014.

\bibitem[Grassia et~al.(2016)Grassia, Kohno, and Levi]{grassia2016digital}
F.~Grassia, T.~Kohno, and T.~Levi, ``Digital hardware implementation of a
  stochastic two-dimensional neuron model,'' \emph{Journal of
  Physiology-Paris}, vol. 110, no.~4, pp. 409--416, 2016.

\bibitem[Gulrajani et~al.(1977)Gulrajani, Roberge, and
  Mathieu]{gulrajani1977modelling}
R.~Gulrajani, F.~Roberge, and P.~Mathieu, ``The modelling of a burst-generating
  neuron with a field-effect transistor analog,'' \emph{Biological
  cybernetics}, vol.~25, no.~4, pp. 227--240, 1977.

\bibitem[Halliday and McDonald(1977)]{halliday1977pathophysiology}
A.~Halliday and W.~McDonald, ``Pathophysiology of demyelinating disease,''
  \emph{British medical bulletin}, vol.~33, no.~1, pp. 21--27, 1977.

\bibitem[Hines et~al.(2009)Hines, Davison, and Muller]{hines2009neuron}
M.~Hines, A.~P. Davison, and E.~Muller, ``Neuron and python,'' \emph{Frontiers
  in neuroinformatics}, vol.~3, p.~1, 2009.

\bibitem[Hodgkin and Huxley(1952)]{hodgkin1952quantitative}
A.~L. Hodgkin and A.~F. Huxley, ``A quantitative description of membrane
  current and its application to conduction and excitation in nerve,''
  \emph{The Journal of physiology}, vol. 117, no.~4, pp. 500--544, 1952.

\bibitem[Huynh et~al.(2022)Huynh, Varshika, Paul, Isik, Balaji, and
  Das]{huynh2022implementing}
P.~K. Huynh, M.~L. Varshika, A.~Paul, M.~Isik, A.~Balaji, and A.~Das,
  ``Implementing spiking neural networks on neuromorphic architectures: A
  review,'' \emph{arXiv preprint arXiv:2202.08897}, 2022.

\bibitem[Ironside(1997)]{ironside1997oppenheimer}
J.~Ironside, ``Oppenheimer's diagnostic neuropathology: A practical manual,''
  \emph{Journal of Clinical Pathology}, vol.~50, no.~2, p. 177, 1997.

\bibitem[I{\c{s}}{\i}k et~al.(2022)I{\c{s}}{\i}k, Paul, Varshika, and
  Das]{icsik2022design}
M.~I{\c{s}}{\i}k, A.~Paul, M.~L. Varshika, and A.~Das, ``A design methodology
  for fault-tolerant computing using astrocyte neural networks,'' \emph{arXiv
  preprint arXiv:2204.02942}, 2022.

\bibitem[Izhikevich(2003)]{izhikevich2003simple}
E.~M. Izhikevich, ``Simple model of spiking neurons,'' \emph{IEEE Transactions
  on neural networks}, vol.~14, no.~6, pp. 1569--1572, 2003.

\bibitem[Joubert et~al.(2012)Joubert, Belhadj, Temam, and
  H{\'e}liot]{joubert2012hardware}
A.~Joubert, B.~Belhadj, O.~Temam, and R.~H{\'e}liot, ``Hardware spiking neurons
  design: Analog or digital?'' in \emph{The 2012 International Joint Conference
  on Neural Networks (IJCNN)}.\hskip 1em plus 0.5em minus 0.4em\relax IEEE,
  2012, pp. 1--5.

\bibitem[Khanday et~al.(2018)Khanday, Kant, Dar, Zulkifli, and
  Psychalinos]{khanday2018low}
F.~A. Khanday, N.~A. Kant, M.~R. Dar, T.~Z.~A. Zulkifli, and C.~Psychalinos,
  ``Low-voltage low-power integrable cmos circuit implementation of integer-and
  fractional--order fitzhugh--nagumo neuron model,'' \emph{IEEE Transactions on
  Neural Networks and Learning Systems}, vol.~30, no.~7, pp. 2108--2122, 2018.

\bibitem[Maeda et~al.(2005)Maeda, Yagi, and Makino]{maeda2005synchronization}
Y.~Maeda, E.~Yagi, and H.~Makino, ``Synchronization with low power consumption
  of hardware models of cardiac cells,'' \emph{BioSystems}, vol.~79, no. 1-3,
  pp. 125--131, 2005.

\bibitem[Meier(2015)]{meier2015mixed}
K.~Meier, ``A mixed-signal universal neuromorphic computing system,'' in
  \emph{2015 IEEE International Electron Devices Meeting (IEDM)}.\hskip 1em
  plus 0.5em minus 0.4em\relax IEEE, 2015, pp. 4--6.

\bibitem[Merolla et~al.(2014)Merolla, Arthur, Alvarez-Icaza, Cassidy, Sawada,
  Akopyan, Jackson, Imam, Guo, Nakamura, et~al.]{merolla2014million}
P.~A. Merolla, J.~V. Arthur, R.~Alvarez-Icaza, A.~S. Cassidy, J.~Sawada,
  F.~Akopyan, B.~L. Jackson, N.~Imam, C.~Guo, Y.~Nakamura \emph{et~al.}, ``A
  million spiking-neuron integrated circuit with a scalable communication
  network and interface,'' \emph{Science}, vol. 345, no. 6197, pp. 668--673,
  2014.

\bibitem[Miller et~al.(2017)Miller, Fyffe-Maricich, and
  Caprariello]{miller2017animal}
R.~H. Miller, S.~Fyffe-Maricich, and A.~C. Caprariello, ``Animal models for the
  study of multiple sclerosis,'' in \emph{Animal models for the study of human
  disease}.\hskip 1em plus 0.5em minus 0.4em\relax Elsevier, 2017, pp.
  967--988.

\bibitem[Moradi et~al.(2017)Moradi, Qiao, Stefanini, and
  Indiveri]{moradi2017scalable}
S.~Moradi, N.~Qiao, F.~Stefanini, and G.~Indiveri, ``A scalable multicore
  architecture with heterogeneous memory structures for dynamic neuromorphic
  asynchronous processors (dynaps),'' \emph{IEEE transactions on biomedical
  circuits and systems}, vol.~12, no.~1, pp. 106--122, 2017.

\bibitem[O'Donnell et~al.(2017)O'Donnell, Gon{\c{c}}alves, Portera-Cailliau,
  and Sejnowski]{o2017beyond}
C.~O'Donnell, J.~T. Gon{\c{c}}alves, C.~Portera-Cailliau, and T.~J. Sejnowski,
  ``Beyond excitation/inhibition imbalance in multidimensional models of neural
  circuit changes in brain disorders,'' \emph{Elife}, vol.~6, p. e26724, 2017.

\bibitem[Sekine(1999)]{sekine1999study}
Y.~Sekine, ``A study on neuronal coding using pulse-type hardware chaotic
  neuron model,'' in \emph{The 3rd International Workshop on Neuronal Coding},
  vol. 121, 1999.

\bibitem[Tsai et~al.(2018)Tsai, Hu, Li, Hwang, and Chou]{tsai2018circuit}
K.-T. Tsai, C.-K. Hu, K.-W. Li, W.-L. Hwang, and Y.-H. Chou, ``Circuit
  variability interacts with excitatory-inhibitory diversity of interneurons to
  regulate network encoding capacity,'' \emph{Scientific reports}, vol.~8,
  no.~1, pp. 1--15, 2018.

\bibitem[Vision(2021)]{vision2021kria}
A.~Vision, ``Kria k26 som: The ideal platform for vision ai at the edge,''
  2021.

\bibitem[Vittoz(1994)]{vittoz1994low}
E.~A. Vittoz, ``Low-power design: Ways to approach the limits,'' in
  \emph{Proceedings of IEEE International Solid-State Circuits
  Conference-ISSCC'94}.\hskip 1em plus 0.5em minus 0.4em\relax IEEE, 1994, pp.
  14--18.

\bibitem[Xue et~al.(2008)Xue, Wang, Deng, and Dong]{xue2008inhibitory}
L.~Xue, J.~Wang, B.~Deng, and F.~Dong, ``Inhibitory chemical coupling of
  electronic morris-lecar neuron model and its bifurcation analysis,'' in
  \emph{2008 30th Annual International Conference of the IEEE Engineering in
  Medicine and Biology Society}.\hskip 1em plus 0.5em minus 0.4em\relax IEEE,
  2008, pp. 2461--2464.

\end{thebibliography}

\end{document}